\documentclass[10pt,twocolumn,letterpaper]{article}

\usepackage{cvpr}              %

\usepackage{graphicx}
\usepackage{amsmath}
\usepackage{amssymb}
\usepackage{booktabs}

\usepackage{times} %
\usepackage{amsmath} %
\usepackage{amssymb}  %
\usepackage{stmaryrd}

\usepackage[utf8]{inputenc}
\usepackage[english]{babel}
\usepackage{adjustbox}
\usepackage{graphicx} %
\usepackage{booktabs} %
\usepackage[table,dvipsnames]{xcolor} %
\usepackage{pifont} %
\usepackage{multirow} %
\usepackage{caption}
\usepackage{subcaption}
\usepackage{sidecap}
\sidecaptionvpos{figure}{t}
\usepackage{flushend}

\usepackage[pagebackref,breaklinks,colorlinks,citecolor=blue]{hyperref}

\newcommand{\tens}{\mathcal}
\newcommand{\mat}{\mathbf}

\newcommand{\acro}{CAB}

\newcommand{\bev}{bird-eye-view}

\def\cvprPaperID{10} %
\def\confName{CVPR}
\def\confYear{2022}

\begin{document}

\title{Raising context awareness in motion forecasting}

\author{H\'edi Ben-Younes\thanks{equal contribution} $^{,1}$, \'Eloi Zablocki\footnotemark[1] $^{,1}$, Micka\"el Chen$^{1}$, Patrick P\'erez$^{1}$, Matthieu Cord$^{1,2}$%
\\\\$^{1}$\,Valeo.ai, $^{2}$\,Sorbonne Université
\\ {\tt \small \{hedi.ben-younes,eloi.zablocki,mickael.chen,patrick.perez,matthieu.cord\}@valeo.com}
}
\maketitle

\begin{abstract}
    Learning-based trajectory prediction models have encountered great success, with the promise of leveraging contextual information in addition to motion history. Yet, we find that state-of-the-art forecasting methods tend to overly rely on the agent's current dynamics, failing to exploit the semantic contextual cues provided at its input. To alleviate this issue, we introduce \acro{}, a motion forecasting model equipped with a training procedure designed to promote the use of semantic contextual information. We also introduce two novel metrics --- dispersion and convergence-to-range --- to measure the temporal consistency of successive forecasts, which we found missing in standard metrics. Our method is evaluated on the widely adopted nuScenes Prediction benchmark as well as on a subset of the most difficult examples from this benchmark. The code is available at \url{github.com/valeoai/CAB}.
\end{abstract}

\section{Introduction}
\label{sec:intro}

\begin{SCfigure*}[0.63]
    \centering 
    \captionsetup{}
    \begin{subfigure}[t]{0.48\linewidth}
        \includegraphics[width=\textwidth]{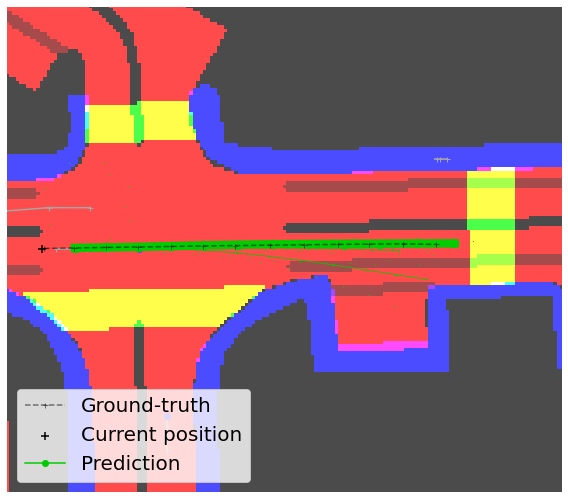}
        \caption{\acro{} (ours)}
    \end{subfigure}
    \hfill
    \begin{subfigure}[t]{0.48\linewidth}
        \includegraphics[width=\textwidth]{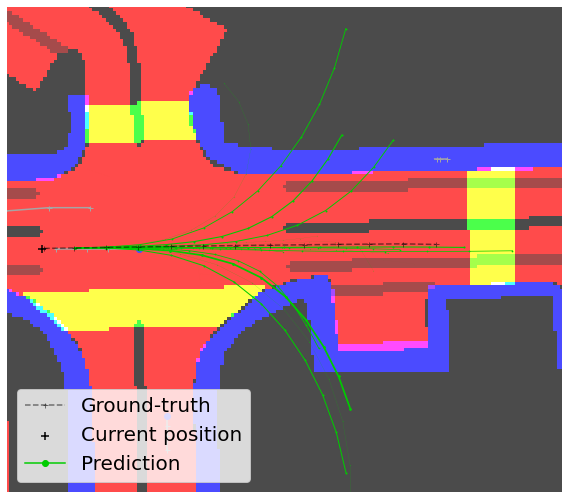}
        \caption{Trajectron++}
    \end{subfigure}        
    \hfill
    \begin{subfigure}[t]{0.48\linewidth}
        \includegraphics[width=\textwidth]{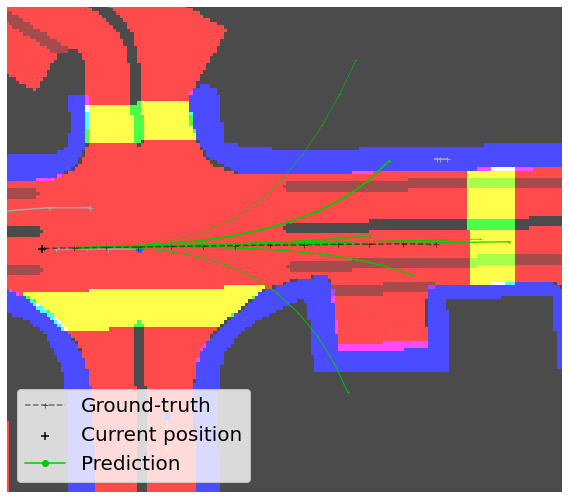}
        \caption{Trajectron++ (no-context)}
    \end{subfigure}
    \caption{Predictions from a) \acro{} (ours), b) Trajectron++, and c) Trajectron++ without the context input.
    The thickness of trajectories represent their likelihood.
    Trajectron++ and its blind variant have very similar predictions and they both forecast trajectories that leave the driveable area. \acro{} is more consistent with the map.
    Sidewalks are in blue, crosswalks in yellow and driveable areas in red.
    }
    \label{fig:big_picture}
    \vspace{-0.2cm}
\end{SCfigure*}

Autonomous systems require an acute understanding of other agents' intention to plan and act safely, and the capacity to accurately forecast the motion of surrounding agents is paramount to achieving this \cite{neural_motion_planner,chauffeurnet,plop}. %
Historically, physics-based approaches have been developed to achieve these forecasts \cite{survey_motion_prediction}.
Over the last years, the paradigm has shifted towards learning-based models \cite{covernet,trajectron++}.
These models generally operate over two sources of information: (1) scene information about the agent's surroundings, \eg LiDAR point clouds \cite{fast_and_furious,intentnet,spagnn,laserflow} or \bev{} rasters \cite{covernet,trajectron++,fast_and_furious,intentnet,trajectron}, and (2) motion cues of the agent, \eg its instantaneous velocity, acceleration, and yaw rate \cite{covernet, mtp} or its previous trajectory \cite{trajectron++,probabilistic_vehicle_trajectory_prediction,mha_jam,multipath}.
But despite being trained with diverse modalities as input, we remark that, in practice, these models tend to base their predictions on only one modality: the previous dynamics of the agent.
Indeed, trajectory forecasting models obtain very similar performances when the scene information about the agent's surroundings is removed from the input (see \autoref{sec:expe}).
This phenomenon stems from the very strong auto-correlations often exhibited in trajectories \cite{nuscenes,argoverse}.
For instance, when a vehicle is driving straight with a constant speed over the last seconds, situations in which the vehicle keeps driving straight with a constant speed are overwhelmingly represented; similarly, if a vehicle starts braking, its future path is very likely a stopping trajectory.
As a consequence, models tend to converge to a local minimum consisting in forecasting motion based on correlations with the past motion cues only, failing to take advantage of the available contextual information \cite{chauffeurnet,off_road_obstacle_avoidance,imitation_learning_end_to_end,exploring_limitations_behavior_cloning_codevilla}.
For example, in \autoref{fig:big_picture}, we observe that several predictions made by the Trajectron++ \cite{trajectron++} model leave the driveable area which hints that the scene information was not correctly used by the model.

Such biased models relying too much on motion correlation and ignoring the scene information are unsatisfactory for several reasons.
First, context holds crucial elements to perform good predictions when the target trajectory is not an extrapolation of the past motion.
Indeed, a biased model will likely fail to forecast high-level behavior changes (\eg start braking), when scene information is especially needed because of some event occurring in the surroundings (\eg front vehicle starts braking).
Leveraging context is thus paramount for motion anticipation, \ie converging quickly and smoothly towards the ground truth ahead in time.
Furthermore, a biased model has a flawed reasoning because it bases its predictions on motion signals rather than the underlying causes contained within the scene environment.
For example, when applied on a vehicle that has started to decelerate, it will attribute its forecast on the past trajectory (\eg `The car will stop because it has started braking.') instead of the underlying reason (\eg `The car will stop because it is approaching an intersection with heavy traffic.') \cite{makansi2022you,explainability_driving_review}.
As a direct consequence, explainability methods analyzing a biased model can lead to less satisfactory justifications.
Overall, it is thus paramount for motion forecasting algorithms to efficiently leverage the contextual information and to ground motion forecasts on it.

In this paper, we propose to equip a motion forecasting model with a novel learning mechanism that encourages predictions to rely more on the scene information, \ie a bird-eye-view map of the surroundings and the relationships with neighboring agents.
Specifically, we introduce \emph{blind} predictions, \ie predictions obtained with past motions of the agent only, without any contextual information.
In contrast, the main model has access to both these inputs but is encouraged to produce motion forecasts that are different from the \emph{blind} predictions, thus promoting the use of contextual information.
Our model is called `\acro{}' as it raises Context Awareness by leveraging Blind predictions.
It is built on the Conditional Variational AutoEncoder (CVAE) framework, widely used in motion forecasting; in practice, it is instantiated with the Trajectron++ \cite{trajectron++} trajectory forecasting backbone. 
Specifically, \acro{} acts on the probabilistic latent representation of the CVAE and encourages the latent distribution for the motion forecasts to be different to the latent distribution for the blind predictions.
Additionally, we introduce Reweight, and RUBiZ, two alternative de-biasing strategies that are not specific to probabilistic forecasting models as they rely on loss and gradients reweighting respectively.

In motion forecast algorithms deployed in real robotic systems, when successive forecasts are done, it is desirable to have both a fast convergence towards the ground-truth, as well as a high consistency of consecutive predictions.
Accordingly, we introduce two novel metrics: \emph{convergence-to-range} and 
\emph{dispersion}.
These metrics aim at providing more refined measurements of how early models are able to anticipate their trajectory and how stable through time their successive predictions are.

Overall, our main contributions are as follows.

1.\,We target the problem of incorporating contextual information into motion forecasting architectures, as we find that state-of-the-art models overly rely on motion. %

2.\,We present \acro{}, an end-to-end learning scheme that leverages blind predictions to promote the use of context. %

3.\,In addition to standard evaluation practices, we propose two novel metrics, namely \emph{dispersion} and \emph{convergence-to-range}, that respectively measure the temporal stability of successive predictions and their spatial convergence speed.

To validate the design of our approach, we conduct experiments on nuScenes \cite{nuscenes}, a public self-driving car dataset focused on urban driving.
We show that we outperform previous works \cite{trajectron++, halentnet}, as well as the alternative debiasing strategies that we propose, inspired by the recent literature in Visual Question Answering (VQA) and Natural Language Inference (NLI) \cite{rubi, nli_bias_mitigation}.
Besides, we use Shapley values to measure the contribution of each modality on the predictions: this allows us to measure how well a model can leverage the context input.
Lastly, we conduct evaluations on a subset of the most difficult examples of nuScenes where we find that our approach is better suited to anticipate high-level behavior changes.

\section{Related Work}
\label{sec:related_work}

Motion forecasting models aim to predict the future trajectories of road agents. This prediction is achieved using information from their current \emph{motion} such as velocity, acceleration or previous trajectory, and some \emph{contextual elements} about the scene. This context can take various forms, ranging from raw LiDAR point clouds \cite{fast_and_furious,intentnet,spagnn,laserflow,desire,r2p2,precog} and RGB camera stream \cite{desire,infer,trafficpredict,sophie} to more semantic representations including High-Definition maps \cite{chauffeurnet,covernet,trajectron++,fast_and_furious,intentnet,trajectron,multi_agent_tensor_fusion,HOME21,GOHOME21,caspnet}, or detections of other agents and their motion information \cite{covernet,trajectron++,mtp,mha_jam}.
Recent trajectory prediction models are designed to produce multiple forecasts, attempting to capture the multiplicity of possible futures \cite{mtp,activity_forecasting,mfp}.
Various learning setups are explored to train these models: regression in the trajectory space \cite{mtp,multipath,rules_of_the_road,convolutional_social_pooling}, spatio-temporal occupancy map prediction \cite{neural_motion_planner, chauffeurnet, HOME21, GOHOME21, caspnet}, or probabilistic methods with either implicit modelling using Generative Adversarial Networks (GANs) \cite{sophie,multi_agent_tensor_fusion,gan,social_gan}, 
 or explicit modelling with Conditional Variational Auto-Encoder (CVAE) \cite{trajectron++,trajectron,desire,precog,cvae}. 
Our work is based on this CVAE family of methods, which not only has provided strong results in motion forecasting but also structurally defines a separation between \emph{high-level decision} and \emph{low-level execution} of this decision \cite{multipath}.

The difficulty of efficiently leveraging contextual information in deep forecasting methods is verified in motion \emph{planning} models that suffer from `causal confusion' on the state variable leading to catastrophic motion drift \cite{off_road_obstacle_avoidance,imitation_learning_end_to_end,exploring_limitations_behavior_cloning_codevilla,causal_confusion}. 
Moreover, models' proclivity to make reasoning shortcuts and to overlook an informative input modality is also encountered in other fields that deal with inputs of different natures, such as medical image processing \cite{medical_image_processing}, Visual Question Answering (VQA), or Natural Language Inference (NLI).
In VQA, for instance, researchers report that models tend to be strongly biased towards the linguistic inputs and mostly ignore the visual input \cite{rubi,analyzing_vqa,dont_just_assume_vqa,overcoming_language_priors_vqa,explicit_bias_discovery_vqa}.
For example, the answer to the question ``What color is the banana in the image'' will be ``Yellow'' 90\% of the time, and models will ignore the image. %
To alleviate this issue, some recent works propose to explicitly capture linguistic biases within a question-only branch and attempt to reduce the impact of these linguistic biases in the general model, for example through adversarial regularization \cite{overcoming_language_priors_vqa}, or with a gradient reweighting strategy during training \cite{rubi}.
We make a parallel between the \emph{current motion} for trajectory forecasting and the linguistic input in VQA.
Also, drawing inspiration from recent de-biasing strategies used in VQA \cite{nli_bias_mitigation, rubi}, we propose novel methods for motion forecasting.
To the best of our knowledge, biases and statistical shortcut on the agent's dynamics have not yet been studied in the context of learning-based motion forecasting.

\section{Model}
\label{sec:model}
The goal is to predict a distribution over possible future trajectories ${\mat{y} = \left[y_1,\ldots,y_T\right]}$ of a moving agent of interest in a scene, where $y_t \in \mathbb{R}^2$ is the position of the agent $t$ steps in the future in a bird-eye view, and $T$ is the prediction horizon.
To do so, we consider a sequence of sensor measurements $\tens{X}$ containing motion information (\eg position, velocity, acceleration) over the $H$ previous steps. Besides, the context $\tens{C}$ provides information about the static (\eg driveable area, crosswalks, etc.) and dynamic (\eg other agents' motion) surroundings of the agent. 
In this framework, a prediction model provides an estimate of $p{\left( \mat{y} | \tens{X}, \tens{C} \right)}$ for any given input pair $(\tens{X}, \tens{C})$.

\begin{figure}
    \centering
    \includegraphics[trim=1.50cm 0.2cm 1.8cm 0.2cm, clip,width=\linewidth]{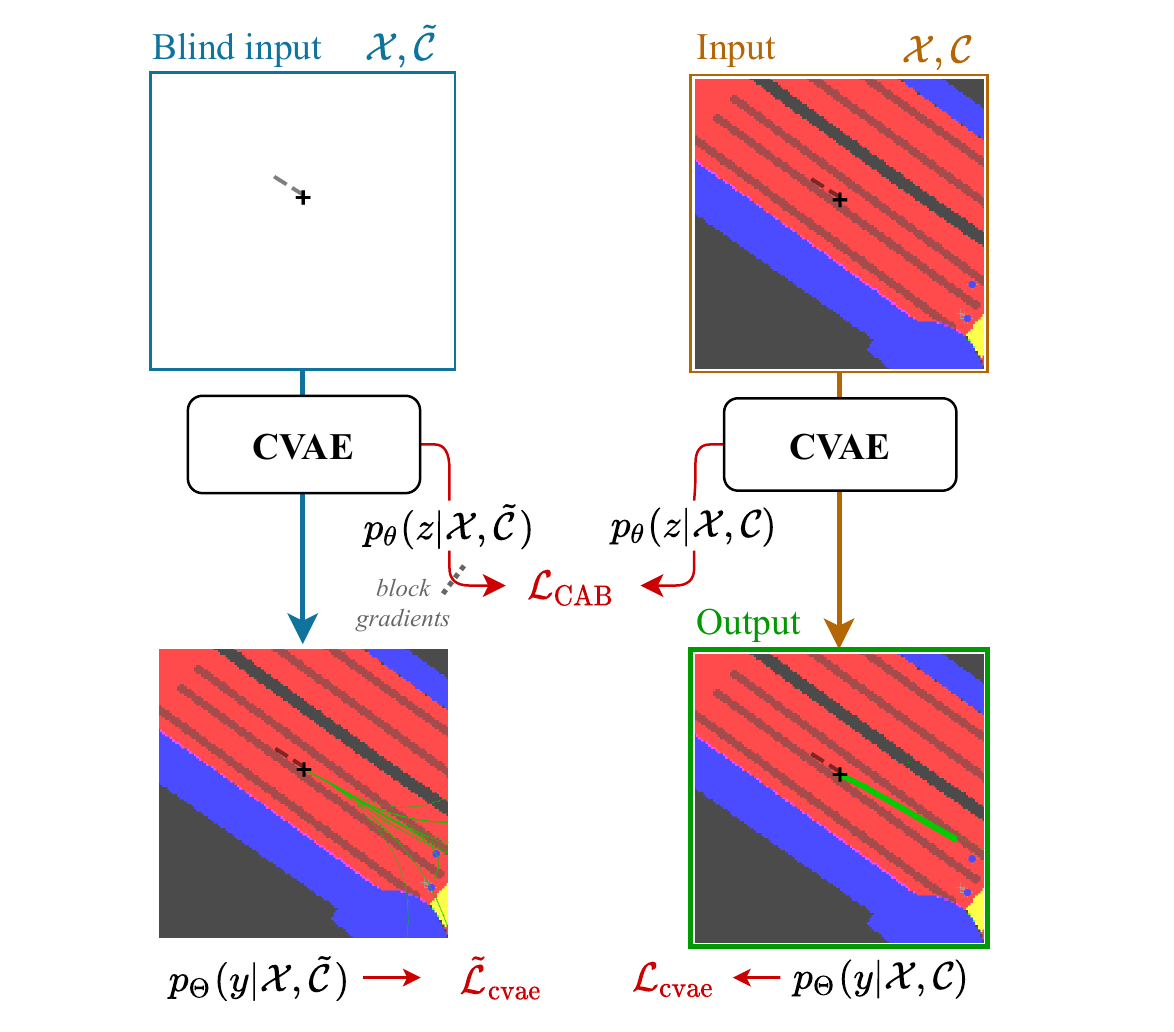}
    \caption{
    \textbf{Overview of the learning scheme of \acro{}.}
    \acro{} employs a CVAE backbone which produces distributions $p_\theta \left(z | \tens{X,C} \right)$ and $p_\Theta{\left(\mat{y} | \tens{X,C} \right)}$ over the latent variable $z$ and the future trajectory.
    During training, a \emph{blind} input $\tens{X,\tilde{C}}$ is forwarded into the CVAE and the resulting distribution over $z$ is used to encourage the prediction of the model to be different from the context-agnostic distribution $p(y|\tens{X})$, thanks to the $\tens{L}_\text{CAB-KL}$ loss.
    Note that the two depicted CVAEs are identical.
    The original context $\tens{C}$ is overlayed onto the prediction for visualization purposes.
    }
    \label{fig:archi}
\end{figure}

\subsection{Conditional VAE framework for motion forecasting}
Following recent works in trajectory forecasting \cite{trajectron++,trajectron,halentnet,desire,precog,mfp}, we use the CVAE framework to train our model for future motion prediction. 
A CVAE provides an estimate $p_\Theta{\left( \mat{y} | \tens{X}, \tens{C} \right)}$ of the distribution of possible trajectories by introducing a latent variable $z \in \tens{Z}$ that accounts for the possible high-level decisions taken by the agent: %
\begin{equation}
\label{eq:y_dist}
    p_\Theta{\left(\mat{y} | \tens{X,C} \right)} = \int_{z \in \tens{Z}} p_\theta \left(z | \tens{X,C} \right) p_\phi{\left(\mat{y} | \tens{X,C},z\right)},
\end{equation}
where $\Theta = \{ \theta, \phi\}$.

To train the CVAE, we need to estimate the latent variable $z$ corresponding to a given trajectory $\mat{y}$. To that end, we introduce the additional distribution $q_\psi {\left( z | \tens{X,C}, \mat{y} \right)}$.

Distributions $p_\theta \left(z | \tens{X,C} \right)$, $p_\phi{\left(\mat{y} | \tens{X,C},z\right)}$ and $q_\psi {\left( z | \tens{X,C}, \mat{y} \right)}$ are parameterized by neural networks, where $\theta$, $\phi$ and $\psi$ are their respective weights. These networks are jointly trained to minimize: 
\begin{equation}
\label{eq:lcvae}
\begin{split}
    \tens{L}_{\text{cvae}} = &\frac{1}{N} \sum_{i=1}^N -\log p_\Theta {\left( \mat{y}_i | \tens{X}_i, \tens{C}_i \right)} \\
    &+ \alpha D_\text{KL} {\large[ q_\psi \left( z | \tens{X}_i, \tens{C}_i, \mat{y}_i \right) \|\, p_\theta \left( z | \tens{X}_i, \tens{C}_i  \right) \large],}
\end{split}
\end{equation}
where the summation ranges over the $N$ training samples indexed by $i$, and $D_\text{KL}$ is the Kullback-Leibler divergence.

\subsection{\acro{}}

Using this setup, ideally, the networks would learn to extract relevant information from both motion and context to produce the most likely distribution over possible outputs $\mat{y}$.
However, because of the very strong correlation between $\tens{X}$ and $\mat{y}$ in driving datasets, they tend, in practice, to learn to focus essentially on $\tens{X}$ and to ignore $\tens{C}$ when estimating $p{\left( \mat{y} | \tens{X}, \tens{C} \right)}$.
In the worse cases, models can collapse into estimating simply $p{\left( \mat{y} | \tens{X} \right)}$.
Yet, $\tens{C}$ contains crucial information such as road boundaries or pedestrians.
Our goal is then to encourage taking $\tens{C}$ into account by introducing a regularization term $\tens{L}_{\text{\acro{}}}$ to the CVAE objective:
\begin{equation}
    \label{eq:loss}
    \tens{L} = \tens{L}_{\text{cvae}} +  \tens{L}_{\text{\acro{}}}.
\end{equation}

The idea of $\tens{L}_{\text{\acro{}}}$ is to encourage the prediction of the model to be different from $p\left( \mat{y} | \tens{X} \right)$.
However, in practice, we do not have access to this distribution.
Instead, we introduce a \emph{blind-mode} for the CVAE model by simply replacing the context input $\tens{C}$ by a \textit{null} context $\tilde{\tens{C}}$.
We obtain $p_\Theta{\large( \mat{y} | \tens{X},\tilde{\tens{C}} \large)}$,
an explicitly flawed model whose predictions can then be used to steer the learning of the main model $p_\Theta{\left( \mat{y} | \tens{X},\tens{C} \right)}$ away from focusing exclusively on $\tens{X}$.

To do so, we would want $\tens{L}_\text{\acro{}}$ to increase $D_\text{KL}
\large[ p_\Theta{\left(\mat{y} | \tens{X,C} \right)}\|\, p_\Theta{(\mat{y} | \tens{X},\tilde{\tens{C}})}\large]$. 
Unfortunately, this term is intractable in the general case, 
and computing a robust Monte-Carlo estimate requires sampling a very large number of trajectories, which would significantly slow down the training.
Therefore, we simplify the problem by setting this divergence constraint on the distributions over $z$ instead of the distributions over $\mat{y}$.
We thus minimize
\begin{equation}
\tens{L}_\text{CAB-KL} = -D_\text{KL} {\large[ p_\theta\left( z | \tens{X,C} \right) \|\, p_\theta ( z | \tens{X}, \tilde{\tens{C}} ) \large]}
\end{equation}
instead.
Following the intuition proposed in \cite{multipath}, the distributions over $z$ model intent uncertainties, whereas distributions over $\mat{y}$ merge intent and control uncertainties. In this case, forcing $p_\theta {\left( z | \tens{X,C}\right)}$ and $p_\theta {( z | \tens{X},\tilde{\tens{C}})}$ to have a high $D_\text{KL}$ explicitly sets this constraint on high-level decisions.

Moreover, to make sure that $p_\Theta{(\mat{y} | \tens{X},\tilde{\tens{C}})}$ is a reasonable approximation for $p{(\mat{y} | \tens{X})}$, we also optimize parameters $\Theta$ for an additional term $\tilde{\tens{L}}_{\text{cvae}}$, which consists in the loss described in \autoref{eq:lcvae} where each $\tens{C}_i$ is replaced by $\tilde{\tens{C}}$.

The final $\tens{L}_{\text{\acro{}}}$ objective is then
\begin{equation}
    \tens{L}_{\text{\acro{}}} = \lambda_{\text{KL}} \tens{L}_\text{CAB-KL} + \lambda \tilde{\tens{L}}_{\text{cvae}},
\end{equation}
where $\lambda$ and $\lambda_{\text{KL}}$ are hyper-parameters.

To ensure that the blind distribution focuses solely on approximating $p(\mat{y} | \tens{X})$, $\tens{L}_{\text{\acro{}-KL}}$ is only back-propagated along $p_\theta ( z | \tens{X}, \tens{C} )$ and not along $p_\theta ( z | \tens{X}, \tilde{\tens{C}} )$. We underline that $\tens{L}_{\text{\acro{}}}$ does not introduce extra parameters. 

\subsection{Instanciation of \acro{} with Trajectron++}
To show the efficiency of \acro{}, we use Trajectron++ \cite{trajectron++}, a popular model for trajectory prediction based on a variant of a CVAE and whose code is freely available.
We first discuss how the loss of Trajectron++ deviates from standard CVAE, and then present its implementation. %

\paragraph{Information Maximizing Categorical CVAE}
Trajectron++ deviates from the standard CVAE setup in two notable ways.
Firstly, following \cite{infovae}, they include in the CVAE objective $\tens{L}_{\text{cvae}}$ a mutual information term $I_q(\tens{X,C}, z)$ between the inputs $\left( \tens{X,C} \right)$ and the latent factor $z$.
Secondly, in Trajectron++, the latent variable $z$ is set as categorical. The output distribution defined in \autoref{eq:y_dist} is then modeled as a Gaussian mixture with $|\tens{Z}|$ modes.
These deviations are easily integrated to \acro{} by adding the same mutual information term and also setting $z$ as categorical. As with Gaussian distributions that are often used in the context of VAEs, the $D_\text{KL}$ between two categorical distributions has a differentiable closed-form expression.

\paragraph{Data and implementation in Trajectron++}
The dynamic history of the agent is a multi-dimensional temporal signal $\tens{X} = [\mat{x}_{\text{--}H}, ..., \mat{x}_{\text{--}1}, \mat{x}_0]$, where each vector ${\mat{x}_j \in \mathbb{R}^8}$ contains position, velocities, acceleration, heading and angular velocity. This sequence is encoded into a vector ${\mat{x} = f_\mat{x} \left( \tens{X} \right)}$, where $f_\mat{x}$ is designed as a recurrent neural network.
The visual context $\tens{C}$ is represented by two quantities that provide external information about the scene. The first is a bird-eye view image $\tens{M} \in \{0,1\}^{h \times w \times l}$, constructed from a high-definition map, where each element $\tens{M}[h,w,l]$ encodes the presence or the absence of the semantic class $l$ at the position $(h,w)$. Classes correspond to semantic types such as ``driveable area'', ``pedestrian crossing'' or ``walkway''. This tensor $\tens{M}$ is processed by a convolutional neural network to provide $\mat{m} = f_{\mat{m}} {\left( \tens{M} \right)} \in \mathbb{R}^{d_m}$.
The second quantity is a vector $\mat{g} \in \mathbb{R}^{d_g}$ that encodes the dynamic state of neighboring agents.
We define the context vector $\mat{c}$ as the concatenation of $\mat{m}$ and $\mat{g}$.

As discussed here-above, distributions $p_\theta \left( z | \tens{X}_i, \tens{C}_i  \right)$ and $q_\psi \left( z | \tens{X}_i, \tens{C}_i, \mat{y}_i \right)$ from \autoref{eq:lcvae} are set as categorical distributions, parameterized by the outputs of neural networks $f_{\theta}(\mat{x},\mat{c})$ and $f_{\psi}(\mat{x},\mat{c},\mat{y})$ respectively.

Then, for each $z \in \tens{Z}$, we have $p_\phi(\mat{y} | \mat{x,c}, z)~=~\tens{N} \left( \mu_z, \Sigma_z\right)$, where $(\mu_z, \Sigma_z) = f_{\phi} \left( \mat{x,c}, z\right)$. These Gaussian densities are weighted by the probabilities of the corresponding $z$, and summed to provide the trajectory distribution:
\begin{equation}
    p_\Theta(\mat{y} | \mat{x,c}) = \sum_{z \in \tens{Z}} p_\theta(z | \mat{x,c}) p_\phi(\mat{y} | \mat{x,c},z).
\end{equation}

Interestingly, $f_\phi$ is constructed as a composition of two functions. The first is a neural network whose output is a distribution over control values for each prediction timestep. The second is a dynamic integration module that models temporal coherence by transforming these control distributions into 2-D position distributions. This design ensures that output trajectories are dynamically feasible.
For more details, please refer to \cite{trajectron++}.

\subsection{Alternative de-biasing strategies}
We also propose two alternative de-biaising strategies.
Like \acro{}, they leverage blind predictions to encourage the model to use the context. However, unlike \acro{} that plays on the specificity of motion forecasting by acting on distribution of the latent representation, these variations are inspired by recent models from the VQA and NLI fields.

$\bullet$ \textbf{Reweight} is inspired by the de-biased focal loss proposed in \cite{nli_bias_mitigation}. The importance of training examples is dynamically modulated during the learning phase to focus more on examples poorly handled by the \emph{blind} model. Formally, the model optimizes the following objective:
\begin{equation}
 \begin{split}
     \tens{L}_{\text{rw}} = &\tens{L}_{\text{cvae}} + \tilde{\tens{L}}_{\text{cvae}} \\
     -&\hspace{-0.07cm} \sum_{i=1}^N \sigma \hspace{-0.0cm} \large(\hspace{-0.00cm} -\log p_\Theta {( \mat{y}_i | \tens{X}_i, \tilde{\tens{C}} )} \hspace{-0.00cm} \large) \log p_\Theta {( \mat{y}_i | \tens{X}_i, \tens{C}_i)},
 \end{split}
\end{equation}
where $\sigma$ represents the sigmoid function.
Intuitively, samples that can be well predicted from the blind model, \ie low value of 
$\sigma\large(-\log p_\Theta ( \mat{y}_i | \tens{X}_i, \tilde{\tens{C}}_i )\large)$, 
will see their contribution lowered and reciprocally, the ones that require contextual information to make accurate forecast, \ie high value of $\sigma\large(-\log p_\Theta ( \mat{y}_i | \tens{X}_i, \tilde{\tens{C}}_i )\large)$, 
have an increased weight.
Similarly to \acro{}, we prevent the gradients to flow back into the blind branch from the loss weight term.

$\bullet$ \textbf{RUBiZ} adjusts gradients instead of sample importance. It does so by modulating the predictions of the main model during training to resemble more to predictions of a blind model. 
RUBiZ is inspired by RUBi \cite{rubi}, a VQA model designed to mitigate language bias. Originally designed for the classification setup, we adapt this de-biasing strategy to operate over the latent factor $z$ of our model, hence the name RUBiZ. In practice, given $\mathbf{l}$ and $\tilde{\mathbf{l}}$ the logits of $p_\theta(z|\tens{X},\tens{C})$ and $p_\theta(z|\tens{X},\tilde{\tens{C}})$, a new distribution over the latent variable is obtained as $p_\theta^{\text{rubiz}}(z|\tens{X}, \tens{C}, \tilde{\tens{C}}) = \text{softmax}(\sigma(\mathbf{l}) * \sigma(\tilde{\mathbf{l}}) )$. This distribution, when used by the decoder, shifts the output of the main prediction towards a blind prediction. Consequently, situations where scene information is essential and past trajectory is not enough have increased gradient, whereas easy examples that are well predicted by the blind model have less importance in the global objective.

\section{Experiments}
\label{sec:expe}

\begin{table*}[t]
    \centering
    \resizebox{\textwidth}{!}{%
        \begin{tabular}{@{\extracolsep{-3pt}}@{}r c c c c c c c c c c c c c c c c c @{}}
            \toprule
            & \multicolumn{6}{c}{ADE-ML} & & \multicolumn{6}{c}{FDE-ML} & OffR-ML & ADE-f & FDE-f & OffR-f \\
            \cline{2-7}
            \cline{9-14}
            \textbf{Model} & @1s & @2s & @3s & @4s & @5s & @6s & & @1s & @2s & @3s & @4s & @5s & @6s & @6s & @6s & @6s & @6s \\
            \hline
            Constant vel.\ and yaw & 0.46 & 0.94 & 1.61 & 2.44 & 3.45 & 4.61 & & 0.64 & 1.74 & 3.37 & 5.53 & 8.16 & 11.21 & 0.14 & - & - & - \\
            \textit{Physics Oracle} & 0.43 & 0.82 & 1.33 & 1.98 & 2.76 & 3.70 & & 0.59 & 1.45 & 2.69 & 4.35 & 6.47 & 9.09 & 0.12 & - & - & - \\
            Covernet, fixed $\epsilon = 2$ & 0.81 & 1.41 & 2.11 & 2.93 & 3.88 & 4.93 & & 1.07 & 2.35 & 3.92 & 5.90 & 8.30 & 10.84 & \textbf{0.11} & - & - & - \\
            Trajectron++ (no-context) & 0.13 & 0.39 & 0.87 & 1.59 & 2.56 & 3.80 & & 0.15 & 0.86 & 2.23 & 4.32 & 7.22 & 10.94 & 0.27 & 4.46 & 12.32 & 0.36 \\
            Trajectron++ & 0.13 & 0.39 & 0.86 & 1.55 & 2.47 & 3.65 & & 0.15 & 0.87 & 2.16 & 4.15 & 6.92 & 10.45 & 0.23 & 4.15 & 11.44 & 0.29 \\
            HalentNet (no-context) & \textbf{0.12} & 0.38 & 0.82 & 1.43 & 2.21 & 3.17 & & \textbf{0.13} & 0.85 & 2.04 & 3.72 & 5.92 & 8.64 & 0.27 & 4.13 & 10.95 & 0.29 \\
            HalentNet & 0.14 & 0.41 & 0.87 & 1.51 & 2.32 & 3.29 & & 0.17 & 0.88 & 2.14 & 3.91 & 6.15 & 8.83 & 0.28 & 3.98 & 10.61 & 0.25 \\
            \hline
            Reweight & 0.13 & 0.38 & 0.81 & 1.42 & 2.20 & 3.14 & & 0.15 & 0.83 & 2.00 & 3.69 & 5.90 & 8.58 & 0.17 & 3.71 & 9.74 & 0.19 \\
            RUBiZ & 0.18 & 0.42 & 0.82 & 1.40 & 2.14 & 3.04 & & 0.23 & 0.84 & 1.95 & 3.55 & 5.65 & 8.21 & \textbf{0.11} & 3.68 & 9.45 & \textbf{0.17} \\
            \acro{} & \textbf{0.12} & \textbf{0.34} & \textbf{0.73} & \textbf{1.29} & \textbf{2.01} & \textbf{2.90} & & 0.14 & \textbf{0.73} & \textbf{1.81} & \textbf{3.39} & \textbf{5.47} & \textbf{8.02} & 0.13 & \textbf{3.41} & \textbf{9.03} & 0.20 \\
            \bottomrule 
        \end{tabular}
    }
    \vspace{-0.1cm}
    \caption{\textbf{Trajectory forecasting on the nuScenes Prediction challenge~\cite{nuscenes}.}
    Reported metrics are the Average/Final Displacement Error (ADE/FDE), and the Off-road Rate (OffR). Each metric is computed for both the most-likely trajectory (\emph{-ML}) and the full distribution (\emph{-f}).
    }
    \label{tab:ml_prediction}
    \vspace{-0.3cm}
\end{table*}

\subsection{nuScenes Dataset}
Our models are trained and evaluated on the driving dataset nuScenes \cite{nuscenes}.
It contains a thousand 20-second urban scenes recorded in Boston and Singapore.
Each scene includes data from several cameras, lidars, and radars, a high-definition map of the scene, as well as annotations for surrounding agents provided at 2 Hz. These annotations are processed to build a trajectory prediction dataset for surrounding agents, and especially for vehicles. Models are trained and evaluated on the official train/val/test splits from the nuScenes Prediction challenge, respectively containing $32186$\,/\,$8560$\,/\,$9041$ instances, each corresponding to a specific agent at a certain time step for which we are given a 2-second history ($H=4$) and are expected to predict up to 6 seconds in the future ($T=12$).

\subsection{Baselines and details}
\noindent\textbf{Physics-based baselines\ \ }
We consider four simple physics-based models, and a \emph{Physics oracle}, as introduced in \cite{covernet}, that are purely based on motion cues and ignore contextual elements.
The four physics-based models use the current velocity, acceleration, and yaw rate and forecast assuming constant speed/acceleration and yaw/yaw rate.
The trajectory predicted by the \emph{Physics oracle} model is constructed by selecting the best trajectory, in terms of average point-wise Euclidean distance, from the pool of trajectories predicted by the four aforementioned physics-based models.
This \emph{Physics Oracle} serves as a coarse upper bound on the best achievable results from a blind model that would be purely based on motion dynamics and ignores the scene structure.

\paragraph{Learning-based forecasting methods}
We compare our debiased models %
against recently published motion prediction models. 
\textbf{CoverNet} \cite{covernet} forwards a rasterized representation of the scene and the vehicle state (velocity, acceleration, yaw rate) into a CNN and learns to predict the future motion as a class, which corresponds to a pre-defined trajectory.
We re-train the ``fixed $\epsilon=2$'' variant, for which the code is available, to compare it with our models.
\textbf{Trajectron++} \cite{trajectron++} is our baseline, which corresponds to removing $\tens{L}_{\acro{}}$ in \acro{}.
\textbf{HalentNet} \cite{halentnet} casts the Trajectron++ model as the generator of a Generative Adversarial Network (GAN) \cite{gan}. A discriminator is trained to distinguish real trajectories from generated ones and to recognize which $z$ was chosen to sample a trajectory. It also introduces `hallucinated' predictions in the training, which correspond to predictions with several confounded values of $z$.
To measure the usefulness of the contextual elements in these models, we also consider the `Trajectron++ (no-context)' and `HalentNet (no-context)' variants that simply discard the map and social interactions from the input of the respective underlying models.
Trajectron++ and HalentNet are not evaluated for different temporal horizons on the nuScenes prediction challenge splits and we thus re-train them given their respective codebases.

\paragraph{Implementation details} We use the ADAM optimizer \cite{adam}, with a learning rate of $0.0003$. The value of hyper-parameters $\lambda = 1.0$ and $\lambda_{\text{KL}} = 5.0$ are found on the validation set.

\subsection{Results and standard evaluations}

We compare our debiased models to the baselines by measuring the widely used metrics of displacement and off-road rate. All models are trained to predict 6 seconds in the future, and their performance is evaluated for varying temporal horizons ($T\in\{2,4,6,8,10,12\}$). 
Average Displacement Error (ADE) and Final Displacement Error (FDE) measure the distance between the predicted and the ground-truth trajectory, either as an average between each corresponding pair of points (ADE), or as the distance between final points (FDE). To compute these metrics with \acro{}, we sample the most likely trajectory $\mat{y}_{\text{ML}}$ by first selecting the most likely latent factor ${z_\text{ML}=\arg\max_{z \in \tens{Z}} p_\theta(z|\mat{x},\mat{c})}$, and then computing the mode of the corresponding Gaussian ${\mat{y}_\text{ML} = \arg\max_y p_\phi(\mat{y}|\mat{x},\mat{c},z_\text{ML})}$. To evaluate the quality of the whole distribution and not just the most-likely trajectory, similarly to \cite{trajectron++,halentnet}, we compute metrics `ADE-f' and `FDE-f'. They are respectively the average and final displacement error averaged for $2000$ trajectories randomly sampled in the full distribution predicted by the network ${\mat{y}_\text{full} \sim p_\Theta(\mat{y}|\mat{x},\mat{c})}$.
Finally, the `off-road rate' (OffR) is the rate of future trajectories that leave the driveable area.

In \autoref{tab:ml_prediction}, we compare the performance of our models \acro{}, Reweight and RUBiZ with baselines from the recent literature. %
To begin with, we remark that for the Trajectron++ model, the use of context brings close to no improvement for predictions up to 4 seconds and a very small one for 5- and 6-second horizons.
Even more surprisingly, the HalentNet (no-context) model which does \emph{not} use any information from the surroundings, shows better ADE-ML and FDE-ML than the regular context-aware HalentNet model. 
This supports our claim that the contextual elements are overlooked by these models and that predictions are mostly done by relying on motion cues.
Moreover, we emphasize that the \emph{Physics oracle} --- which is purely based on motion dynamics --- obtains very strong performances (3.70 ADE-ML@6s, 9.09 FDE-ML@6s) as it can choose the closest trajectory to the ground truth from a variety of dynamics.
Its scores approximate upper bounds on the forecasting abilities of purely motion-based models and we observe that learning-based methods hardly outperform this \emph{Physics-oracle} on long temporal horizons.

On the other hand, we remark that all three of our debiaising strategies significantly outperform the \emph{Physics oracle} and previous models on almost all the metrics, both when looking at the most-likely trajectory as well as the full future distribution.
This validates the idea, shared in our methods, to enforce the model's prediction to have a high divergence with a blind prediction.
Indeed, despite optimizing very different objective functions, our \emph{Reweight} and \emph{RUBiZ} and \acro{} share the idea of a motion-only encoding.
More precisely, at a 6-second horizon, the sample reweighting strategy gives a relative improvement of 16\% w.r.t. Trajectron++. %
The more refined RUBiZ strategy of gradient reweighting gives a relative improvement of 19\% w.r.t. Trajectron++. %
\acro{} achieves a 22\% relative improvement over Trajectron++. This indicates that guiding the model's latent variable constitutes a better use of \emph{blind} predictions than simple example or gradient weightings.

\subsection{Further analyses: stability, convergence, Shapley values}
\label{sub:further_eval}

\begin{figure*}
    \centering
    \includegraphics[width=\linewidth]{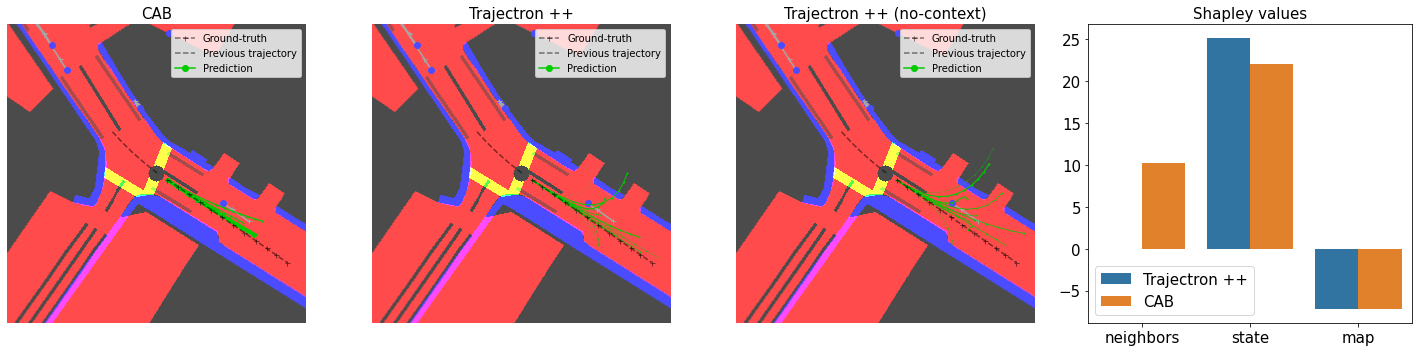}
    \vspace{0.2cm}
    \includegraphics[width=\linewidth]{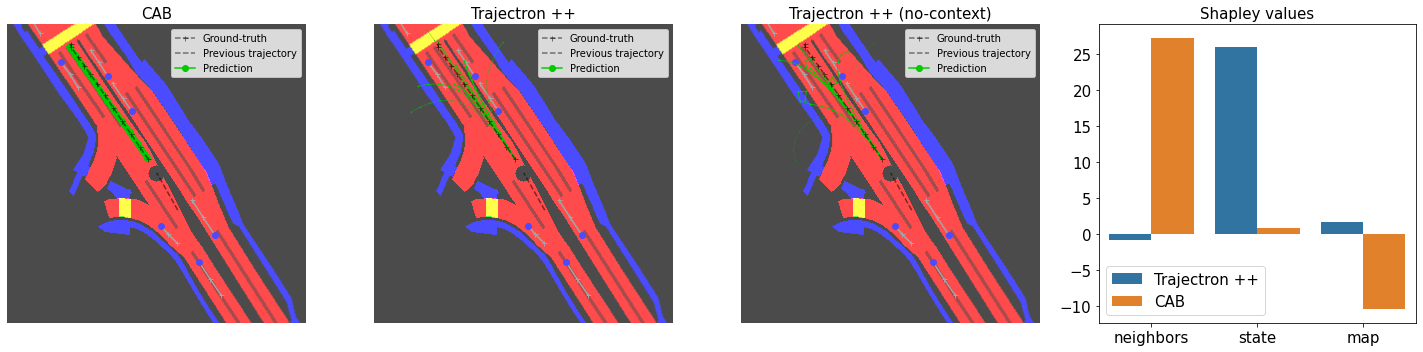}
    \caption{\vspace{-0.1cm}\textbf{Visualizations of predicted trajectories and Shapley values}. The thickness of lines represent the probability of each trajectory.} %
    \label{fig:visus_shapley}
\end{figure*}

We hypothesize that properly leveraging contextual information has a strong impact on the ability to anticipate the agent's intents. %
Intuitively, for an agent arriving at an intersection, a model without context will begin predicting a stopping trajectory only from the moment when this agent starts to stop, whereas a model with a proper understanding of contextual information will be able to foresee this behavior change ahead in time.
Furthermore, improving this anticipation ability should also help the temporal stability of the predictions, as unanticipated changes of trajectory will be less likely.
Unfortunately, ADE and FDE metrics do not explicitly measure the rate of convergence towards the ground-truth, nor the stability of successive predictions. %

\begin{table}[t]
    \centering
    \resizebox{\columnwidth}{!}{%
        \begin{tabular}{@{}rcccc@{}}
            \toprule
            & \multirow{2}{*}{Dispersion $D$ $\downarrow$}& \multicolumn{3}{c}{Convergence-to-range $C(\tau)$ $\uparrow$} \\
            \textbf{Model} & & $\tau = 20$cm & $\tau = 1$m & $\tau = 5$m \\
            \hline
            Cst.\ accel., yaw & 6.55 & 0.44 & 1.49 & 3.30 \\
            Cst.\ accel., yaw rate & 6.03 & 0.47 & 1.58 & 3.46 \\
            Cst.\ speed, yaw rate & 3.42 & 0.54 & 1.82 & 4.15 \\
            Cst.\ vel., yaw & 3.33 & 0.53 & 1.82 & 4.16 \\
            \textit{Physics Oracle} & 2.99 & 0.55 & 1.91 & 4.38 \\
            \hline
            Covernet, fixed $\epsilon = 2$ & 4.06 & 0.15 & 0.93 & 3.50 \\
            Trajectron++ (no-context)  & 3.65 & 1.00 & 2.11 & 4.17 \\
            Trajectron++ &  3.55 & 0.98 & 2.11 & 4.22 \\
            Halentnet (no-context) & 2.85 & 1.03 & 2.28 & 4.59 \\
            Halentnet  &  3.23 & 0.90 & 2.07 & 4.26 \\
            \hline
            Reweight & 3.02 & 1.03 & 2.27 & 4.50 \\
            RUBiZ & 2.73 & 0.94 & 2.31 & 4.60 \\
            \acro{} & \textbf{2.61} & \textbf{1.12} & \textbf{2.45} & \textbf{4.74} \\
            \bottomrule 
        \end{tabular}
    }
    \vspace{-0.1cm}
    \caption{\textbf{Study of the temporal stability of trajectory prediction}, with the Dispersion ($D$) and Convergence-to-range ($C(\tau)$) metrics. Predictions are made at a 6-second horizon.} %
    \label{tab:stability}
    \vspace*{-0.7em}
\end{table}

Consequently, we introduce two new metrics focusing on the stability and convergence rate of successive forecasts.
Instead of classically looking at the $T$-step forecast $\mat{y}^t = {\left[ y^t_1, \dots, y^t_T \right]}$ made at a specific time $t$, we take a dual viewpoint by considering the consecutive predictions $\large[ y_T^{\hat{t}-T}, \dots, y_1^{\hat{t}-1} \large]$ made for the same ground-truth point $y^{\text{gt}}_{\hat{t}}$.
When the agent approaches the timestamp $\hat{t}$, as $t$ grows, predictions $y_t^{\hat{t}-t}$ will get closer to the ground-truth $y^{\text{gt}}_{\hat{t}}$.
In addition to low ADE/FDE scores, it is desirable to have both (1) a high consistency of consecutive predictions, as well as (2) a fast convergence towards the ground-truth $y^{\text{gt}}_{\hat{t}}$.
Therefore, for a given annotated point at $\hat{t}$, we define the \emph{dispersion} $D_{\hat{t}}$ as the standard deviation of the points predicted by the model for this specific ground-truth point $D_{\hat{t}}~=~\texttt{STD}~\large(~\| y^{\hat{t}-t}_t~-~\bar{y}_{\hat{t}}\| \large)_{t\in\llbracket 1,T\rrbracket}$ where $\bar{y}_{\hat{t}}$ is the barycenter of $\{ y^{\hat{t}-t}_t \}_{t\in\llbracket 1,T\rrbracket}$.
The global dispersion score $D$ is obtained by averaging these values over all the points in the dataset.
Moreover, we propose the \emph{convergence}-to-range-$\tau$ metric $C(\tau)_{\hat{t}}$ as the time from which all subsequent predictions fall within a margin $\tau$ of the ground-truth $y^{\text{gt}}_{\hat{t}}$, where $\tau$ is a user-defined threshold:
\begin{equation}
    C_{\hat{t}}(\tau)\!=\!\max \big\{ T'\!\in\llbracket 1, T \rrbracket \!\mid\! \forall t \leq T', \| y^{\hat{t}-t}_t\!-\! y^{\textit{gt}}_{\hat{t}} \|_2 \leq \tau \big\}.
\end{equation}

In \autoref{tab:stability}, we report evaluations of the stability and spatial convergence metrics.
First, we observe that previous learning-based forecasting models have more limited anticipation capacities than the simpler physics-based models, in terms of both convergence speed (metric $C(\tau)$) and convergence stability (metric $D$).
Consistently to the results of \autoref{tab:ml_prediction}, we remark that our de-biased strategies, and especially our main model \acro{}, lead to better anticipation scores as they converge faster towards the ground truth.

In \autoref{fig:visus_shapley}, we visualize trajectories generated by \acro{} and the baselines Trajectron++ and Trajectron++ (no-context). 
We also analyze the contribution brought by each input of the model.
To do so, we estimate the Shapley values \cite{shap} which correspond to the signed contribution of individual input features on a scalar output, the distance to the final predicted point in our case. %
We remark that the Shapley value of the state signal is overwhelmingly higher than the ones attributed to the map and the neighbors for Trajectron++. This means that the decisions are largely made from the agent's dynamics. This can further be seen as several predicted trajectories are highly unlikely futures as they collide with the other agents and/or leave the driveable area.
Instead, the Shapley values for \acro{} give much more importance to both the map and the neighboring agents, which helps to generate likely and acceptable futures.

\vspace*{-1em}
\subsection{Evaluation on hard situations}

\begin{table}[t]
    \centering
    \resizebox{\columnwidth}{!}{%
        \begin{tabular}{@{}r c c c c@{}}
            \toprule
             \textbf{Model} & 1\% & 2\% & 3\% & All\\
             \hline 
             Trajectron++ (no-context) & 15.80 & 15.58 & 14.97 & 10.94 \\
             Trajectron ++ & 13.12 & 12.69 & 12.25 & 10.45 \\
             HalentNet & 14.12 & 12.83 & 12.09 & 8.83 \\
             \hline
             Reweight & 14.00 & 13.30 & 12.58 & 8.58\\
             RUBiZ & 13.42 & 12.49 & 11.64 & 8.21 \\
             \acro{} & \textbf{12.13} & \textbf{11.88} & \textbf{11.59} & \textbf{8.02} \\
            \bottomrule
        \end{tabular}
    }
    \vspace{-0.1cm}
    \caption{
    \textbf{Final Displacement Error FDE@6s on challenging situations,} as defined by Makansi et al. \cite{challenging_long_tail_future_prediction}. Results for columns `$i$\%' are averaged over the top $i$\% hardest situations as measured by the the mismatch between the prediction from a Kalman filter and the ground-truth. \vspace{-0.3cm}
    }
    \label{tab:hard_examples}
    \vspace{-0.1cm}
\end{table}

\begin{figure}[t]
    \centering
    \includegraphics[width=0.49\linewidth]{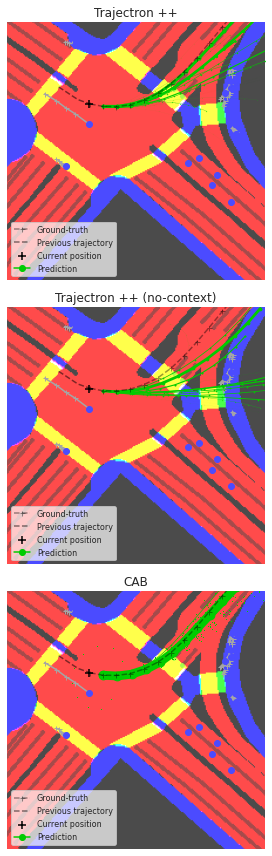}
    \hfill
    \includegraphics[width=0.49\linewidth]{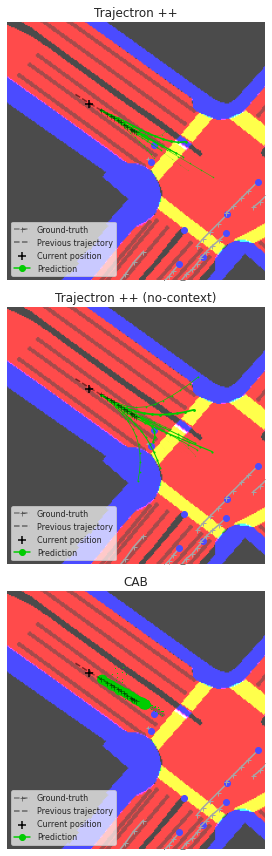}
    \caption{
    \textbf{Visualizations on challenging situations,} as defined by Makansi et al. \cite{challenging_long_tail_future_prediction}. %
    By better leveraging the context, \acro{} generates more accurate predictions while Trajectron++ leaves the driveable area or collides into other agents.
    } 
    \label{fig:visus_hard_examples}
    \vspace*{-1.5em}
\end{figure}

We verify that the performance boost observed in \autoref{tab:ml_prediction} does not come at the expense of a performance drop on difficult yet critical situations.
Accordingly, we use recently proposed evaluations \cite{challenging_long_tail_future_prediction} as they remark that uncritical cases dominate the prediction and that 
complex scenarios cases are at the long tail of the dataset distribution.
In practice, situations are ranked based on how well the forecast made by a Kalman filter fits the ground-truth trajectory. 

In \autoref{tab:hard_examples}, we report such stratified evaluations, on the 1\%, 2\%, and 3\% hardest situations.
Our first observation is that the overall performance (`All') does not necessarily correlate with the capacity to anticipate hard situations.
Indeed, while HalentNet significantly outperfoms Trajectron++ on average, it falls short on the most challenging cases.
Besides, \acro{} achieves better results than Trajectron++ on the hardest situations (top 1\%, 2\%, and 3\%).
Lastly, while the gap between Trajectron++ and CAB is only 0.66 point for the 3\% of hard examples, it increases for the top 1\% of hardest examples up to 0.99.

In \autoref{fig:visus_hard_examples}, we display some qualitative results we obtain on challenging situations selected among the 1\% hardest examples.
On the left, we observe that the turn is not correctly predicted by Trajectron++ as it estimates several possible futures that leave the driveable area.
On the right, the agent of interest has to stop because of stopped agents in front of it and this behavior is well forecasted by \acro{}, unlike Trajectron++ which extrapolates the past and provides multiple futures colliding into other agents.
Overall, the better use of the context in \acro{} not only helps on average situations but also on difficult and potentially critical ones.

\section{Conclusion}

We showed that modern motion forecasting models struggle to use contextual scene information.
To address this, we introduced \emph{blind} predictions that we leveraged with novel de-biaising strategies.
This results into three motion forecasting models designed to focus more on context.
We show that doing so helps reducing statistical biases from which learning-based approaches suffer. In particular, \acro{}, which is specifically built for probabilistic forecasting models, makes significant improvements in traditional distance-based metrics.
Finally, after introducing new stability and convergence metrics, we show that \acro{} shows better anticipation properties than concurrent methods.

\vspace*{0.cm}{\noindent\textbf{Acknowledgments}: We thank Thibault Buhet, Auguste Lehuger and Ivan Novikov for insightful comments. This work was supported by ANR grant VISA DEEP (ANR-20-CHIA-0022) and MultiTrans (ANR-21-CE23-0032).

{\small
\bibliographystyle{ieee_fullname}
\bibliography{biblio_cleaned}
}

\end{document}